# Comparing Open Arabic Named Entity Recognition Tools

**Abdullah Aldumaykhi\*, Saad Otai\*, Abdulkareem Alsudais**
College of Computer Engineering and Sciences, Prince Sattam bin Abdulaziz University
Al-Kharj 11942, Saudi Arabia
{aldomaikhi, 441540018, a.alsudais}@psau.edu.sa

**Abstract**
The main objective of this paper is to compare and evaluate the performances of three open Arabic NER tools: CAMeL, Hatmi, and Stanza. We collected a corpus consisting of 30 articles written in MSA and manually annotated all the entities of the person, organization, and location types at the article (document) level. Our results suggest a similarity between Stanza and Hatmi with the latter receiving the highest F1 score for the three entity types. However, CAMeL achieved the highest precision values for names of people and organizations. Following this, we implemented a "merge" method that combined the results from the three tools and a "vote" method that tagged named entities only when two of the three identified them as entities. Our results showed that merging achieved the highest overall F1 scores. Moreover, merging had the highest recall values while voting had the highest precision values for the three entity types. This indicates that merging is more suitable when recall is desired, while voting is optimal when precision is required. Finally, we collected a corpus of 21,635 articles related to COVID-19 and applied the merge and vote methods. Our analysis demonstrates the tradeoff between precision and recall for the two methods.

## 1. Introduction

Named entity recognition (NER) is the task of identifying known entities such as people and organizations in text data. Methods for NER rely on predefined rules or lists, statistical information, machine learning techniques, and recently, large pretrained language models. NER can be used to enhance other natural language processing (NLP) and information retrieval applications such as question answering (Phan V and Do, 2021; Alwaneen et al., 2021), machine translation (Li et al., 2020; Ugawa et al., 2018), and semantics (Alsudais and Tchalian, 2019; Khare et al., 2018). Three of the most commonly used types of named entities are person (PER), organization (ORG), and location (LOC). Although the performance of computational models that automatically identify named entities in text data has improved in recent years, NER remains a research challenge. Moreover, research has traditionally focused on NER at the sentence level and limited research has focused on NER at the article or document levels (Wang et al., 2018). For the Arabic language, several NER methods have been introduced over the years. A handful have ready-to-use tools or software that allow users to find named entities in documents written in Arabic. In this paper, our primary objective is to collect and annotate a new dataset that consists of Arabic documents and to compare the performance of three open Arabic NER tools as well as two of our own proposed approaches that integrate these three. Our sample is a leading Saudi newspaper, providing a specific focus on how these tools process names that are more common in the country.

Arabic is the official language in several countries in the Middle East and North Africa. It is the native language for over 300 million individuals and is considered one of the most morphologically diverse natural languages in the world. Some of the unique characteristics of Arabic, such as its lack of orthographic standards (Habash et al., 2018; Harrat et al., 2019), its diverse dialects (Darwish et al., 2021; Alsudais et al., 2022), and the limited existing NLP resources make Arabic NLP challenging. There are also problems that make Arabic NER particularly difficult. Shaalan (2014) described many of the challenges, including Arabic's "lack of capitalization" and "lack of uniformity in writing styles." Despite this, several methods for Arabic NER currently exist. Some of these methods include tools that can be implemented and tested by interested individuals.

Three tools that have recently been particularly successful for Arabic NER are Stanza, CAMeL, and hatmimoha/arabic-ner (Hatmi). Stanza is a natural language processing toolkit that supports several languages, including Arabic. CAMeL tools is an open-source toolkit that is focused on supporting tasks in Arabic NLP, including an Arabic named entity tagger. Hatmi is a pre-trained BERT-based NER model for Arabic that relies on HuggingFace's Transformers (Wolf et al., 2020). The three vary in their scope and the methods they use to identify Arabic named entities. To the best of our knowledge, no previous studies have examined these three tools and compared their performances when used on the same corpus at the article-level. This gap in the scholarship represents a main motivation for our work. More specifically, we are comparing the performance of these three tools on a dataset of Arabic news articles that were published in a leading Saudi newspaper. We chose to select a dataset that would provide a focus on this one country in order to investigate the ideal NER tool to use when the analyzed corpus included a higher number of named entities that are more common in Saudi. Additionally, we have tested "merging" and "voting" methods that utilize the results from these three tools. Finally, we have collected a subset of news articles containing the term "Covid-19" and discuss various observations about the use of these five options to identify named entities.

A study that focuses on comparing these three current NER tools can have several benefits. First, our comparative study identifies the method that performs best when used to identify named entities of the Person, Organization, and Location types. In order to furnish a dataset with many possible named entities, we collected a large dataset of Arabic news articles from a leading Saudi newspaper, then selected a random sample of 30 articles that we tagged manually. This new dataset represents a second benefit of our study: 30 articles with all the persons, organizations, and locations in each article manually identified and tagged at the article level. It was unclear to us whether combining the results obtained from the three models could generate results that were more accurate than using each model alone. Therefore, we tested a merge method that combines all the results obtained from the models independently and a vote method that only tags named entities when two of the three models recognize it. Finally, we collected a large dataset of news articles that contained mentions of COVID-19 and its related terms. Our analysis of this subset sheds light on how the five methods perform when used on a new dataset with many data and terms that are novel.

---

\*Abdullah Aldumaykhi and Saad Otai contributed equally.

In summary, we made the following key contributions in this paper:

- We manually annotated a sample of 30 articles and tagged all named entities of the Person, Organization, and Location types at the document level.
- We tested the performances of three open NER tools and quantified their results. This represents the first comparison of these three tools on the same corpus.
- We improved the results by testing two approaches: 1) a merging method that combines all the entities extracted from the three tools and 2) a voting method that only accepts a tagged NE when at least two of the methods identify it as a named entity. This represents a major contribution of our work: Results for these two proposed methods outperform the results obtained when these models are used independently.
- We collected a dataset of 21,635 articles written about COVID-19 and published in a leading Saudi newspaper.
- We discussed several observations made when we used the five methods to extract named entities from this new corpus.

## 2. Related Work

Arabic Named Entity Recognition has been a research challenge for many years. Researchers have used various methods and techniques in an attempt to improve Arabic NER. Traditional approaches to the problem have relied on rule-based models (Zaghouani, 2012; Shaalan and Raza, 2009; Oudah and Shaalan, 2016; Abdallah et al., 2012). These approaches were followed by models that were based on machine learning (Darwish, 2013; Benajiba et al., 2010; Abdelali et al., 2016), as well as by hybrid approaches (Oudah and Shaalan, 2012; Shaalan and Oudah, 2014). Recently, several authors have experimented with deep learning techniques (Helwe and Elbassuoni, 2019; Liu, 2019; Al-smadi et al., 2020; Helwe et al., 2020; Khalifa and Shaalan, 2019). Researchers developing NER models have been using annotated corpora (i.e., corpora that include articles or sentences, with tagged entities in each article or sentence). There are several annotated corpora for Arabic NER. One popular dataset is AQMAR (Mohit et al., 2012), which comprises 28 articles from Wikipedia with tagged named entities of the Person, Organization, Location, and Miscellaneous types. This corpus has often been used as a benchmark for new NER models (Liu, 2019; Helwe and Elbassuoni, 2019; Helwe et al., 2020). Another commonly references corpus is ANERcorp (Benajiba et al., 2007b), which contains entities of the Person, Organization, and Location types. Both AQMAR and ANERcorp contain text written in Modern Standard Arabic. Alternatively, CANERCorpus is a corpus that contains documents written in Classical Arabic (Salah and Binti Zakaria, 2018), in which named entities are also tagged. Some additional corpora have targeted specific domains. For example, one study focused on the medical domain and manually annotated medical diseases that appeared in 27 medical articles (Alshammari and Alanazi, 2020). While most of these previous corpora have seemed to focus on Modern Standard Arabic, some did develop corpora and models for Classical Arabic (Alsaaran and Alrabiah, 2021) and Dialectical Arabic (Zirikly and Diab, 2014; Torjmen and Haddar, 2020; Attia et al., 2018). In this paper, we are focusing on a corpus written in Modern Standard Arabic written for a Saudi newspaper.

One of the contributions of this paper is a new corpus that consists of articles that reference COVID-19. Currently, several other Arabic COVID-19 corpora exist. Many of these corpora focused on information posted on Twitter. First, Haouari et. al. (Haouari et al., 2021a) collected 2.7M Arabic tweets about the pandemic that were posted in the period between January the 27th of 2020 till January the 31st of 2021. The same authors also published a datasets focusing on rumors (Haouari et al., 2021b). Another Twitter dataset focused on misinformation and misleading information and included both Arabic and English tweets (Elhadad et al., 2021). A similar dataset also focused on the same topic and included 10K tweets that were manually annotated (Hadj Ameur and Aliane, 2021). Other Twitter datasets also exist (Alqurashi et al., 2020; Alsudias and Rayson, 2021). To the best of our knowledge, one current gap is the lack of datasets that focused on news stories about COVID-19 that were published in Arabic newspapers.

## 3. Methods

In this section, we describe the methodological steps we followed in this study. First, we report our data collection process and explain the data annotation approach. Then, we detail the three methods we used and explain their underlying structure. Following this, we describe the "merge" and "vote" methods we are testing and outline the evaluation criteria. Finally, we describe the "COVID-19" subset and the objectives of using it with the five methods.

### 3.1 Dataset

Our first step was to collect a dataset of Arabic news articles. In this study, we wanted to focus on articles published in a Saudi newspaper. To the best of our knowledge, no previous studies have focused on such a dataset. Due to the prevalence of certain names of people, organizations, and locations, we believed it was possible that one of the methods tested might perform better when used on such a subset. To collect the subset, we focused on Al-Riyadh newspaper, a leading Saudi newspaper. For each collected article, we captured relevant information such as the article's ID, its publication date, title, and textual contents. We noticed that the publication dates were written in several formats. Therefore, we implemented a process to unify all publication dates. Since NER is sensitive to preprocessing steps that may remove stop words and punctuations, we did not perform any preprocessing for the textual contents of articles. In the end, we had a total of

242,724 IDs of articles that were published in the newspaper between 2017 and 2021. From these, we selected 1) a random sample of 30 that we manually annotated and 2) a subset of 21,635 articles that contained terms related to COVID-19.

## 3.2 Annotation

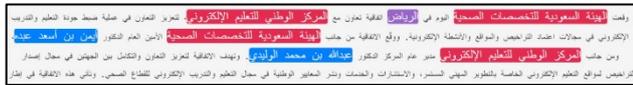

Figure 1: A sample sentence that shows the manual annotation process.

For the 30 randomly selected articles, we identified all named entities of the Person (PER), Organization (ORG), and Location (LOC) types. The number of articles in our annotated corpus is similar to the sizes of corpora used in other studies for Arabic NLP. To facilitate the annotation process, we used DataTurks annotation tool (DataTurks, 2020). One important distinction in our annotation process is that we counted named entities at the article level. Put differently, if the same organization is mentioned twice in the same article, our annotation only counted it once. We decided to follow this annotation process because our subsequent analysis using the COVID-19 subset relies on counting named entities at the article level, a common counting strategy when analyzing large corpora. Such a strategy has the advantage of not overcalculating certain entities that appeared many times in the same article. Moreover, we believe this focus provide novel findings since our annotations does not focus on sentence-level annotation of named entities. Two of the authors of this paper completed the annotation, which was then confirmed by the third author. To ensure agreement on what constitutes a named entity of the PER, ORG, and LOC types, we selected an initial sample that contained only 10 articles. The two authors responsible for annotation completed the labeling process. Then, all three authors discussed the findings and confirmed a uniform understanding of the definitions (i.e. when to tag a word or phrase as a location). Following this step, we revisited the 30 articles, and the two annotators completed the labeling, which was then discussed and confirmed with the third author. We used this annotated sample of 30 articles to evaluate the various methods. Figure 1 shows an example of labeling of named entities using DataTurks. Entities of the PER type are in blue, ORG are in red, and LOC are in purple. For example, the city "Riyadh الرياض" is present and is thus tagged in purple. In the end, our annotation resulted in identifying 54 named entities of the PER type, 64 of ORG, and 88 of LOC.

## Tools and Methods

Following the data collection and annotation process, we wanted to test the five approaches. In this section, we will describe each method. The first three are tools that have been provided by the scientific community. In our selection process, we focused on tools that are freely available and can be easily accessed. The rationale behind this selection condition is to only test tools and methods that are accessible and can be efficiently implemented by users. Moreover, we selected only tools that relied on methods that have been scientifically published.

### 3.2.1 Stanza

The first software library we tested was Stanza. It is a natural language processing toolkit created by the Stanford NLP Group (Qi et al., 2020). The library is written in Python and includes support for several NLP tasks such as tokenizing, sentiment analysis, and NER. Moreover, the library supports multiple languages such as English, Spanish, and Arabic. The Arabic NER model was evaluated on the AQMAR (Mohit et al., 2012) corpus. The authors of Stanza indicated that their NER model relied on the utilization of "a forward and a backward character-level LSTM language model", word embeddings, and a "standard one-layer Bi-LSTM sequence tagger with a conditional random field (CRF)-based decoder" (Qi et al., 2020). Stanza is an extension of Stanford CoreNLP and is available as a Python package. In this study, we downloaded this package and implemented its Arabic NER model.

### 3.2.2 CAMeL

The second software library we used was CAMeL tools (Obeid et al., 2020). It includes a set of functions for a variety of NLP tasks such as sentiment analysis, Arabic dialect classification, and NER. One main difference between CAMeL and Stanza is that the former is specifically designed for and only supports Arabic. CAMeL is also available as a Python package. The software was developed by researchers from the CAMeL lab in New York University Abu Dhabi. The NER model in CAMeL was based on using "HugginFace's Transformers (Wolf et al., 2020) to fine-tune AraBERT" (Antoun et al., 2020). According to the authors of CAMeL, their NER model was evaluated on the ANERcorp (Benajiba et al., 2007a) and the overall F1 result was 83% while it was 90% for LOC, 63% for MISC, 73% for ORG, and 87% for PER. The model supports these four types of named entities.

### 3.2.3 Hatmi

Hatmi is a pre-trained BERT-based NER model for Arabic created by Mohamed Hatmi. Although there is no peer-reviewed paper that explains and evaluates the model, its author indicated that when tested on a "valid corpus made of 30.000 tokens [Hatmi] shows an F-measure of ~87%" (Mohamed Hatmi, 2020). Additionally, the author indicated that the model is based on Arabic-BERT-base (ArabicBERT) (Safaya et al., 2020), which was pretrained on around 8.2 billion words. The model is available via HugginFace and supports the following entities (classes): person, organization, location, date, product, competition, prize, event, and disease. For the three models, we completed initial testing to ensure they met our requirements and that preliminary results would be promising when tested on our corpus.

| Type | Annotated Entities | CAMeL | Stanza | Hatmi | Merge | Vote |
|---|---|---|---|---|---|---|
| PER | ['عمر بن محمد الفريح'] | ['عمر بن محمد الفريح'] | ['عمر بن محمد الفريح'] | ['عمر بن محمد الفريح'] | ['عمر بن محمد الفريح'] | ['عمر بن محمد الفريح'] |
| ORG | ['فيسبوك'] | ['فيسبوك'] | NaN | NaN | ['فيسبوك'] | NaN |
| LOC | ['نيويورك', 'ويست فيرجينيا', 'الولايات المتحدة', 'كاليفورنيا'] | ['نيويورك', 'ويست فيرجينيا', 'الولايات المتحدة', 'كاليفورنيا'] | ['ويست فيرجينيا', ' كاليفورنيا', ' نيويورك', 'كورونا المستجد', ' جامعة جونز هوبكنز', ' الولايات المتحدة'] | ['نيويورك', 'ويست فيرجينيا', 'الولايات المتحدة', 'كاليفورنيا'] | ['كاليفورنيا', 'جامعة جونز هوبكنز', 'الولايات المتحدة', 'ويست فيرجينيا', 'كورونا المستجد', 'نيويورك'] | ['نيويورك', 'ويست فيرجينيا', 'الولايات المتحدة', 'كاليفورنيا'] |

Table 1: Examples of how the merge and vote methods determines named entities

### 3.2.4 Merge Method

The fourth approach we tested was to merge all the results from the previous three tools. When examining the initial results from a sample of articles, we noticed that the three tools did not always identify the same set of entities. This was the main motivation for testing the merge method. It works by combining the results from the models into one list. For example, for a sample paragraph, each model might identify the same organization. However, in some cases, we noticed that each model recognized a different organization. Thus, we inferred that combining all the results might improve performance. To implement this method, we wrote a function that combined all discovered named entities but captured the same entity only once. Table 1 includes a sample of manually annotated named entities from several sentences and the results reached when we used the "merge" method. We used "NaN" when one method did not identify any NEs of the specified type. For example, the first row shows only one entity of the "PER" type and all three were able to recognize it. In the third row, we only labeled four entities, but combining all those discovered from the three methods resulted in a list of six entities. Two of these six were inaccurately tagged as named entities of the Location type. Thus, this was an indicator that this method may not actually improve results.

### 3.2.5 Vote Method

One issue with merging all the results from the three models is that any mistakenly identified entity by any of the three will be included in the results for the merge method. Therefore, we wanted to also test a voting method. This method identifies a named entity only when two or more of the three tools identify it as such. Since we observed that many mistakes or false positives only occurred in one of the three models, we tested this method to quantify whether it reduced the number of these false positives. Additionally, we wanted to examine if implementing this method would eliminate what could be the merge method's strongest advantage: its ability to capture all the existing named entities. Table 1 includes the same three examples of the manually labeled entities as well as the results obtained using the vote method. As the results show, in row 2, this method eliminated the correct organization which was only identified by CAMeL. However, in the final row, the two named entities inaccurately captured by the merge method were not identified as locations by the vote method. These initial results provided justifications for the experimentation on and quantification of the performances of the five approaches.

### 3.3 Evaluation

To evaluate the performances of the five approaches, we used three standard metrics: precision, recall, and F1 measures. The three are commonly used to evaluate models that identify named entities in text (Ahmad et al., 2020; Khalifa and Shaalan, 2019; Kanwal et al., 2019). Calculating the three relied on identifying True Positive (TP), False Positive (FP), and False Negative (FN) cases. Using our annotated lists of named entities from the 30 articles, we measured how each method performed based on its TP, FP, and FN numbers. We defined TP instances as those where we annotated a named entity that the model also identified and tagged as such. For FP, this took place when the model identified an entity that we did not annotate as an entity. Finally, FN were instances when we annotated an entity that the method did not identify. It is important to note here that for each method, we calculated these values for the three types of entities (PER, ORG, and LOC) and then for the three combined (four total). Once these values were calculated, we used them to obtain the results for the precision, recall, and F1 measures. The formulas we used were as follows:

- Precision = TP/TP+FP. In our case, precision was the ratio of correctly identified named entities divided by the total number of entities identified as named entities by the model (both correctly and incorrectly).
- Recall = TP/TP+FN. Recall represented the ratio of correctly identified positive NEs divided by the total number of manually annotated named entities.
- F1 Measure = 2 * (Recall * Precision) / (Recall + Precision). Finally, the F1 measure represented the weighted average of precision and recall

### 3.4 Analyzing a "COVID-19" Corpus

The final step in our work was to analyze the performances of the methods when used on a COVID-19 corpus. The objective here was to examine how these methods performed when applied to a collection of documents that included discussions of a developing story that was likely not seen when training these methods. To collect the COVID-19 corpus, we amassed 21,635 articles published in the same newspaper. All the articles had one or more of the following words: "coronavirus," "corona," "covid" as well as "corona" and "covid" when written in Arabic. The sixth and final word was the Arabic word for "pandemic" which is "جائحة". We included this word because we observed that it was frequently used in the newspaper to describe the COVID-19 pandemic.

# 4. Results

In this section, we detail the results we obtained for the five methods. First, we describe the results for Stanza, CAMeL, and Hatmi when used independently. Then, we show the results for the merge and vote methods and compare them to the other three methods. Finally, we apply the methods with the highest performances on the COVID-19 corpus and discuss some of our observations.

## 4.1 Results for Stanza, CAMeL, and Hatmi

| NE Type | Model | Precision | Recall | F1 Scores |
|---|---|---|---|---|
| PER | Stanza | 0.648148 | **0.648148** | 0.648148 |
| | CAMeL | **0.833333** | 0.092593 | 0.166667 |
| | Hatmi | 0.717391 | 0.611111 | **0.66** |
| ORG | Stanza | 0.277778 | 0.15625 | 0.2 |
| | CAMeL | **0.318182** | 0.109375 | 0.162791 |
| | Hatmi | 0.264706 | **0.28125** | **0.272727** |
| LOC | Stanza | 0.39604 | **0.454545** | 0.42328 |
| | CAMeL | 0.421053 | 0.181818 | 0.253968 |
| | Hatmi | **0.5** | 0.375 | **0.428571** |
| ALL | Stanza | 0.445026 | **0.412621** | 0.428212 |
| | CAMeL | 0.424242 | 0.135922 | 0.205882 |
| | Hatmi | **0.466667** | 0.407767 | **0.435233** |

Table 2: Summary of the results for Stanza, CAMeL, and Hatmi

The results for the three methods varied across the three entity types. The detailed results are in Table 2. For each entity type in the table, we highlighted in bold the method with the highest performance for each metric. For example, CAMeL had the highest precision value for PER with 0.83 compared to 0.64 and 0.71 for Stanza and Hatmi, respectively. Results show CAMeL receiving high precision and low recall values for all three entity types. However, precision values were still higher for Stanza and Hatmi when results for the three entity types were combined. Precision and recall results for Stanza and Hatmi were similar, and Hatmi had slightly higher values for precision while Stanza had higher recall values. Additionally, when all the entity types were combined, F1 scores showed that Hatmi had the highest performances compared to the other two methods. This suggests that when given the option to select only one of the three to identify named entities in similar corpora, Hatmi should be selected. Despite this, the F1 score result for Stanza was similar. Thus, existing frameworks that rely on that library may not benefit from switching to Hatmi. Finally, F1 scores for CAMeL were significantly lower, presumably due to CAMeL's lower recall values. Still, the tool scored highest on some of the metrics. For example, it had the highest precision value when identifying names of people. In the end, the varying performances and clear strengths and weaknesses of each tool prompted us to consider combining the results from the three.

## 4.2 Results for the Merge and Vote Methods

| NE Type | Model | Precision | Recall | F1 Scores |
|---|---|---|---|---|
| PER | Merge | 0.578947 | **0.814815** | **0.676923** |
| | Vote | **0.964286** | 0.5 | 0.658537 |
| ORG | Merge | 0.264957 | **0.484375** | **0.342541** |
| | Vote | **0.444444** | 0.0625 | 0.109589 |
| LOC | Merge | 0.358209 | **0.545455** | **0.432432** |
| | Vote | **0.58** | 0.329545 | 0.42029 |
| ALL | Merge | 0.376147 | **0.597087** | **0.461538** |
| | Vote | **0.689655** | 0.291262 | 0.409556 |

Table 3: Summary of the results for the merge and vote methods

Following our individual evaluation of each tool, we combined all the results from the three and used the identified named entities to evaluate the results for the two Results for the merge and vote methods for the different entity types show a similar pattern. For the three and including the total (all), the merge method scored highest for recall and F1 scores, while the vote method performed best for precision. This result represents another major finding of this work: Using the merge or vote methods will yield better performances than using Stanza, CAMeL, or Hatmi individually. The results for the merge and vote methods are in Table 3. Values in bold represent the highest scores when the five methods were compared. For example, F1 scores for the merge method when used on PER were higher than those for Stanza, CAMeL, Hatmi, and the vote method. Overall, voting between the three models generated higher precision values. This is perhaps because voting reduced false positive cases that only one of the three identified as a named entity. In the end, these results indicate that when precision is most desirable, voting does produce the best values while merging has the highest overall scores for F1 and recall. Therefore, selecting either of these two methods might depend on the particular use case and application. These results also highlight some of the differences between the approaches. For example, the recall value for voting when used on ORG was very low. This suggests that there were few true positive instances of ORG that two of the three models agreed on. Figure 2 compares the five approaches.

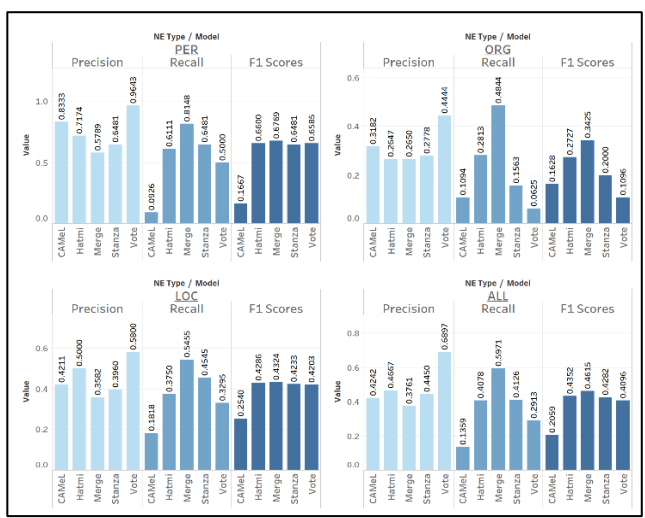

Figure 2: Comparison of the performances of the five methods

## 4.3 Analyzing the COVID-19 Corpus

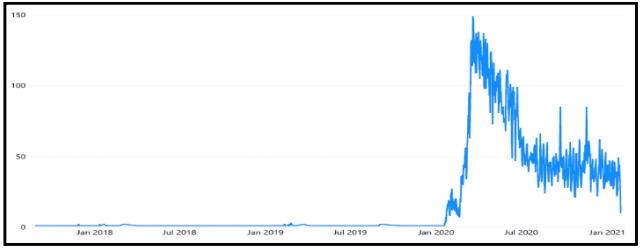

Figure 3: The number of articles in the COVID-19 corpus

After determining which method worked best, we wanted to apply the merge and vote methods to a new, large corpus, specifically the 21,635 articles from a leading Saudi newspaper that contained reference to the COVID-19 pandemic. First, we wanted to see the number of articles we had for the time period we specified (see Figure 3). We noticed that there were a few articles that were published before 2020. Upon inspection, we found that these articles were related to the Middle Eastern Respiratory Syndrome, a species of coronavirus that affected countries in the Middle East before 2020. It is important to note that the number of these articles was limited compared to the overall number of articles. Following this discovery, we applied the merge and vote method on the corpus and ranked the most frequent PERs, ORGs, and LOCs.

The results illustrated the tradeoff between precision and recall for the merge and vote methods. While the merge methods identified a significantly larger number of named entities, many of the names that appeared to be the most frequent were incorrectly identified as named entities. For example, "covid" and "corona" were two of the most frequent named entities of the person type found using the merge method. Interestingly, "corona" was also the most frequent named entity for ORG and LOC. Additionally, four of the most frequent LOCs were COVID-19 related terms that included "corona (كورونا)," "coronavirus (فايروس كورونا)," "covid (كوفيد)," and "novel corona (كورونا المستجد)." Nevertheless, the remaining entities on the lists of the ten most frequent NEs were identified as 1) accurate names of entities that were 2) involved in or affected by the pandemic. These included organizations such as the World Health Organization, the Saudi Ministry of Health, and the United Nations. In the end, one option to improve the results obtained using the merge method is to use expert judgements. Put differently, an expert can examine the lists of most frequent named entities and remove those that are not correct names of entities. For the vote method, we observed that none of the ten most frequent PERs, ORGs, or LOCs (30 total) were inaccurate. This represented an additional demonstration of the value of the high precision results for the vote method. Moreover, none of the "COVID-19" related terms appeared as a top PER, ORG, or LOC. Additionally, we observed that the city "Riyadh" was identified as both a top organization and a location using the merge method, while voting only listed it as a top location. Still, voting does suffer from low recall, which results in a failure to identify many of the named entities that it should capture. For example, the voting method identified 545 articles with the ORG "World Health Organization" while the merge method identified 875. Table 4 includes the ten most frequent named entities of the ORG type that we identified using the merge and vote methods.

| Merge | | Vote | |
|---|---|---|---|
| Term | Freq | Term | Freq |
| كورونا | 3274 | منظمة الصحة العالمية | 545 |
| وزارة الصحة | 1470 | لمنظمة الصحة العالمية | 148 |
| المملكة | 1448 | الهلال | 142 |
| منظمة الصحة العالمية | 875 | وزارة الصحة الكويتية | 124 |
| مجلس الوزراء | 591 | ريال مدريد | 87 |
| المملكة العربية السعودية | 549 | وزارة الصحة | 82 |
| الرياض | 403 | النصر | 81 |
| تويتر | 394 | الاتحاد الأوروبي | 79 |
| الاتحاد الأوروبي | 321 | وزارة الصحة العمانية | 78 |
| للأمم المتحدة | 293 | الحكومة البريطانية | 77 |

Table 4: The most frequent organizations found using the merge and vote methods

The results also highlighted some of the difficulties associated with Arabic named entity recognition in general. First, we observed that the names of non-Arabic individuals and organizations were often written using two or more spellings. For example, the names "Donald Trump" appeared twice (two different spellings) in the list of the 10 most frequent PERs. Second, because some prepositions in Arabic join the words they precede, most frequent lists included the names of organizations twice: once with a preposition and once without. For example, the "World Health Organization" was once listed as that ("منظمة الصحة العالمية") and once with the preposition "to" ("لمنظمة الصحة العالمية"). Finally, we noticed instances where the same name was written either with or without the inclusion of a middle name. Similarly, we observed that sometimes the names of individuals were written with the Arabic word "son of" ("بن") before the middle (father) name. In all of these examples, it appeared as though the two names were different when they in fact belonged to the same real-life entity. In the end, these grammatical and spelling conventions in Arabic represent some of the several challenges that face Arabic named entity recognition. Some of which might be specific or more common in newspaper articles from Saudi Arabia.

## 5. Conclusions

In this paper, we annotated a collection of 30 articles and tagged all the named entities of the PER, ORG, and LOC types. We then tested three popular Arabic NER tools and evaluated their results. Moreover, we experimented with combining the results from the three tools using merge and vote methods. Our findings indicated that the vote method yielded the highest precision values while the merge method generated the highest recall and F1 measure values. These are the key findings of our work. We applied the merge and vote methods on a corpus of articles about COVID-19. We discussed several observations and examined the list of most frequent named entities.

Despite the utility of this work, it also has several potential limitations. First, our selection criteria for tools focused on those that have accessible software implementations and avoided others that were only discussed in papers without having a tool that could be downloaded and utilized. Including these other papers could have resulted in the selection of models with better performances than those we selected. Second, our annotated corpus included articles written in a Saudi newspaper, and thus, our results may not be generalizable to all documents written in Modern Standard Arabic. For example, many of the named entities of the PER type were Saudi individuals. It is possible that one of the models does not capture such names. While we believe that these limitations are negligible, they still represent opportunities for future research which could focus on additional models and investigate whether results for the merge and vote methods change. Farasa (Abdelali et al., 2016) is one such tool that could be useful for future work. Our approach could also be tested on other corpora. In the end, we believe our work presents several key contributions that help provide a better understanding of the issue of Arabic NER at the article-level and the possibilities associated with testing a variety of its tools and models.

## 6. Bibliographical References